\documentclass[10pt,conference]{IEEEtran}

\usepackage{amsmath}
\usepackage{amssymb}
\usepackage{amsfonts}
\usepackage{graphicx}

\usepackage{algorithm}

\usepackage{algpseudocode}

\usepackage{xcolor}

\usepackage{cite}

\algblockdefx{Repeat}{EndRepeat}[1]{\textbf{repeat} #1}{\textbf{end repeat}}

\algdef{SE}[DOWHILE]{Do}{doWhile}{\algorithmicdo}[1]{\algorithmicwhile\ #1}%

\usepackage[margin=0.7in]{geometry}
\usepackage{setspace}

\title{Robust Distributed Bayesian Learning with Stragglers via Consensus Monte Carlo \vspace{-0.4cm}}

\author{
\IEEEauthorblockN{Hari Hara Suthan Chittoor, ~ Osvaldo Simeone \\}
\IEEEauthorblockA{
KCLIP Lab, Department of Engineering, Kings College
London, UK. \\
Email: \{hari.hara, osvaldo.simeone\}@kcl.ac.uk}
\vspace{-0.89cm}
}

\begin{document}

\date{\today}
\maketitle

\thispagestyle{empty}	

\pagestyle{empty}


{\let\thefootnote\relax\footnotetext
{The authors have received funding from  the European Research Council(ERC) under the European Unions Horizon 2020 Research and Innovation Programme (Grant Agreement No. 725731).
}
}

\begin{abstract}

This paper studies distributed Bayesian learning in a setting encompassing a central server and multiple workers by focusing on the problem of mitigating the impact of stragglers. The standard one-shot, or embarrassingly parallel, Bayesian learning protocol known as consensus Monte Carlo (CMC) is generalized by proposing two straggler-resilient solutions based on grouping and coding. Two main challenges in designing straggler-resilient algorithms for CMC are the need to estimate the statistics of the workers' outputs across multiple shots, and the joint non-linear post-processing of the outputs of the workers carried out at the server. This is in stark contrast to other distributed settings like gradient coding, which only require the per-shot sum of the workers' outputs. The proposed methods, referred to as Group-based CMC (G-CMC) and Coded CMC (C-CMC),  leverage redundant computing at the workers in order to enable the estimation of global posterior samples at the server based on partial outputs from the workers. Simulation results show that C-CMC may outperform G-CMC for a small number of workers, while G-CMC is generally preferable for a larger number of workers.

\end{abstract}

\begin{IEEEkeywords}
Distributed Bayesian learning, stragglers, Consensus Monte Carlo, grouping, coded computing
\end{IEEEkeywords}


\section{Introduction}
One of the main problems in  distributed computing systems \cite{A_Berkeley_View_of_Systems_Challenges_for_AI,scaling_distributed_Machine_Learning_with_the_Parameter_Server,mapreduce_J_Dean_2008,TensorFlow:_Large-Scale_Machine_Learning_on_Heterogeneous_Distributed_Systems} is the presence of stragglers -- i.e., working machines whose random computing time is much larger than other machines \cite{The_Tail_at_Scale}. The effect of stragglers may be mitigated by leveraging redundant storage and computing at the workers, whereby each worker is allocated, and computes over, multiple data  shards. State-of-the-art techniques leverage \emph{grouping}, whereby groups of workers are assigned the same shards and compute the same output, and/or \emph{coding}, whereby computed outputs are coded at the workers and jointly decoded at the server  \cite{GC_Rashish_Tandon,LAGC,Kannan_Speeding_Up_Distributed_Machine_Learning_Using_Codes}.

Existing work on grouping and coded distributed computing for machine learning applications focuses on frequentist learning. In \emph{frequentist learning}, the goal is to identify a single model parameter vector that approximately minimizes the training loss, e.g., via gradient descent \cite{GC_Rashish_Tandon,LAGC,Kannan_Speeding_Up_Distributed_Machine_Learning_Using_Codes}. Frequentist learning is limited in its ability to quantify uncertainty, incorporate prior knowledge, guide active learning, and enable continual learning. \emph{Bayesian learning} provides a principled approach to address all these limitations, at the cost of an increase in computational complexity  \cite{OnCalibrationofModernNeuralNetworks,Simple_and_Scalable_Predictive_Uncertainty_Estimation_Using_Deep_Ensembles,Book_Patterns_of_Scalable_Bayesian_Inference,book_Bayesian_Reasoning_and_Machine_Learning,Book_Osvaldo}.

 
\begin{figure}
    \centering
    \includegraphics[height=2.15in]{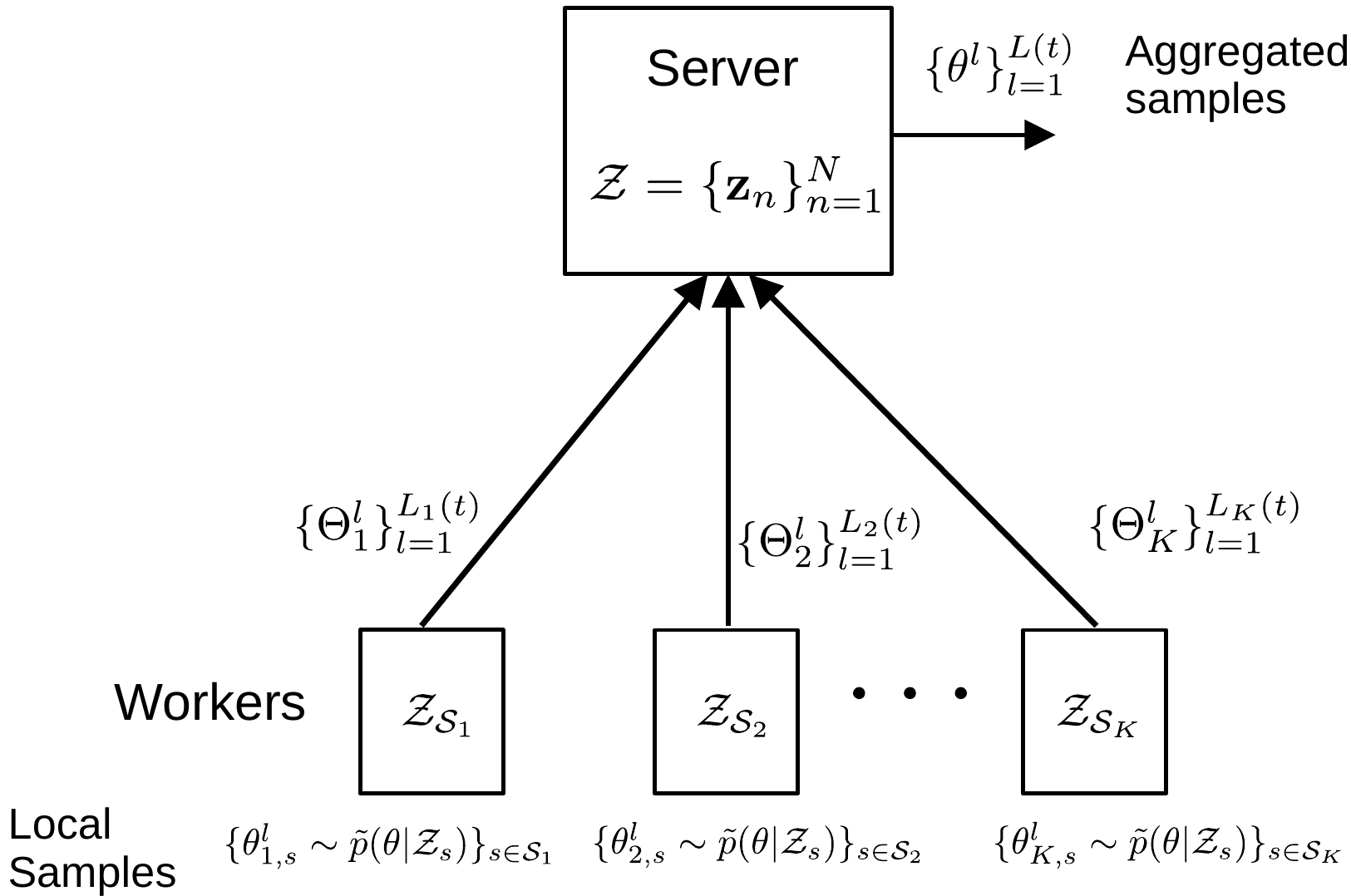} \vspace{-0.2cm}
    \caption{Distributed Bayesian learning via Consensus Monte Carlo (CMC).}
    \label{fig:GCMC}
    \vspace{-0.8cm}
\end{figure} 


Scalable implementations of Bayesian learning are based on either  \textit{variational inference (VI)} -- replacing integration with optimization over an approximate posterior distributions --  or \textit{Monte Carlo (MC) sampling} -- replacing integration with sampling from the posterior distribution \cite{Book_Patterns_of_Scalable_Bayesian_Inference}. VI-based protocols for distributed Bayesian learning follow the same general principles of standard distributed frequentist learning (e.g., \cite{bui2018partitioned} and references therein). Distributed MC sampling protocols are either one-shot,  i.e.,  embarrassingly parallel  \cite{CMC_Bayes_and_Bigdata,Variational_CMC,Rahif_DSVGD}; or else based on iterative gradient-based methods \cite{Distributed_Stochastic_Gradient_MCMC,Dongzhu_Langevin} .
    
In this paper, we focus on the standard one-shot protocol known as \emph{Consensus Monte Carlo (CMC)}  \cite{CMC_Bayes_and_Bigdata}. CMC aims at obtaining samples from the global posterior distribution based on local sampling at the workers and aggregation at the server (see Fig.  1). CMC assumes that all workers respond to the server by delivering their local samples before a global sample can be produced by the server. This paper considers, for the first time, the problem of stragglers for CMC. 
    
Two extensions of CMC are proposed that obtain resilience to stragglers based on grouping and coding. The first protocol, referred to as \emph{Group-based CMC (G-CMC)}, requires the partition of workers into groups, with each group being responsible for the computation of local samples for a given subset of shards. In contrast to grouping methods proposed for frequentist learning and distributed computing  \cite{Grouping_in_distributed_Gradient_descent,LAGC}, a novel feature of G-CMC scheme is that all the computed samples can be eventually utilized, even if produced by straggling workers. The second protocol, \emph{Coded-CMC (C-CMC)} applies an erasure correcting code to the produced samples, in a manner similar to gradient coding \cite{GC_Rashish_Tandon}. Unlike gradient coding, C-CMC requires the design of a novel  pre-processing step of the local samples in order to enable CMC-based aggregation at the server.
    





\section{System Model}
\label{section system model}

\vspace{-0.3cm}

This paper considers the problem of drawing samples from a posterior distribution on a large training data set to implement Bayesian learning via Monte Carlo (MC) sampling. Let $\mathcal{Z}=\{\mathbf{z}_n\}_{n=1}^{N}$ represents the training data and $\theta \in \mathbb{R}^d$ represents the model parameter vector. The global posterior distribution is given as  \vspace{-0.3cm}
\begin{equation}
\label{eqn global posterior}
    \text{(Global Posterior)~~~~ } p(\theta | \mathcal{Z}) \propto p(\theta) p(\mathcal{Z} | \theta),
\end{equation}
where $p(\theta)$ is the prior distribution and $p(\mathcal{Z}| \theta )$ is the likelihood. We assume that the data points are conditionally independent and identically distributed (i.i.d.), given the model parameter vector $\theta$, i.e., $p(\mathcal{Z} | \theta) = \prod_{n=1}^{N} p(\mathbf{z}_n | \theta)$. The goal is to draw $L$ samples $\{\theta^l\}_{l=1}^{L}$ from the global posterior $p(\theta |\mathcal{Z})$. 

We adopt a data center computing platform that consists of a server and $K$ workers, as shown in Fig. \ref{fig:GCMC}. The training data $\mathcal{Z}$ is partitioned into $K$ disjoint shards $\mathcal{Z}=\{\mathcal{Z}_s\}_{s=1}^{K}$, each of size $N/K$ where $K$ assumed to be an integer divisor of $N$, and allocated to the workers by following a data allocation scheme. 
We allow for a  redundant shard allocation, so that each shard is allocated to $r$ workers, where $r \in [K]\triangleq \{1,...,K\}$ is referred to as the redundancy parameter. 
Each worker $k$ has a set $\mathcal{S}_k \subseteq [K]$ of $|\mathcal{S}_k|=r$ shards, which are denoted as $\mathcal{Z}_{\mathcal{S}_k} = \{\mathcal{Z}_s\}_{s \in \mathcal{S}_k}$, with per worker storage capacity $Nr/K$.
Following CMC, we assume that the sampling from each subposterior, \vspace{-0.1cm}
\begin{equation}
\label{eqn subposterior}
    \text{(Subposterior)~~~~} \tilde{p}(\theta | \mathcal{Z}_s) \propto p(\theta)^{^{1/K}} p(\mathcal{Z}_s| \theta),
\end{equation}
is tractable, where $p(\mathcal{Z}_s | \theta) = \prod_{\mathbf{z}\in \mathcal{Z}_s} p(\mathbf{z} | \theta)$ represents the local likelihood function for the $s$-th shard. 
In (\ref{eqn subposterior}), the prior is underweighted as $p(\theta)^{1/K}$ in order to preserve the total prior, so that the global posterior (\ref{eqn global posterior}) can be expressed as the product of subposteriors $p(\theta | \mathcal{Z}) \propto \prod_{s=1}^{K} \tilde{p}(\theta | \mathcal{Z}_s)$.

Each worker $k$ computes in parallel $r$ samples $\theta_{k,s}^{l} \sim \tilde{p}(\theta | \mathcal{Z}_s)$ for all the $r$ allocated shards $\mathcal{Z}_{s:~ s\in \mathcal{S}_k}$, where index $l\in \{1,2,\cdots\}$ runs over the generated samples. We refer to the collection of such samples as $\Theta_{k}^{l} = \{\theta_{k,s}^{l}\}_{s\in \mathcal{S}_k}$. 
The produced samples may be processed at the worker, and the outcome of this calculation is sent to the server. The server uses this information to produce global samples $\theta^l$, that are approximately distributed according to the global posterior distribution (\ref{eqn global posterior}).

We assume that the wall-clock time $\Delta T_k^l$ required to compute any $l$-th batch $\Theta_k^l$ of $r$ local samples from the subposteriors (\ref{eqn subposterior}) of the shards allocated to each worker $k$ is random with mean $\eta r$, for some $\eta >0$. The computing times $\{\Delta T_k^l\}_{k \in [K]}$ are i.i.d. across the workers and across index $l$ \cite{GC_Rashish_Tandon,scaling_distributed_Machine_Learning_with_the_Parameter_Server,LAGC}. An example distribution of the computing time is Pareto with scale-shape parameters $(\eta r(\beta - 1)/ \beta, \beta)$, which gives mean $\eta r$, with $\eta >0, \beta >1$ as constants \cite{LAGC}.

For any continuous time $t$, define as $L(t)$ the number of global samples produced at the server based on the information received so far from the workers. Following the prior works \cite{Variational_CMC,Dongzhu}, we evaluate the error of a CMC algorithm by fixing a test function $f(\cdot)$, and comparing the empirical average obtained with the produced global samples, $\{\theta^l\}_{l=1}^{L(t)}$, available at time $t$ with the corresponding ensemble average $\mathrm{E}_{p(\theta | \mathcal{Z})} [f(\theta)]$ with respect to the true posterior distribution $p(\theta | \mathcal{Z})$ in (\ref{eqn global posterior}). This can be written as \vspace{-0.3cm}
\begin{equation}
\label{eqn error}
\mathrm{err}(t)= \dfrac{\Big|\frac{1}{L(t)}\sum_{l=1}^{L(t)} f(\theta^{l}) - \mathrm{E}_{p(\theta | \mathcal{Z})} [f(\theta)] \Big|}{\mathrm{E}_{p(\theta | \mathcal{Z})} [f(\theta)]}.
\end{equation}

\vspace{-0.3cm}

\begin{figure}[htbp]
    \centering
    \includegraphics[height=2.6in]{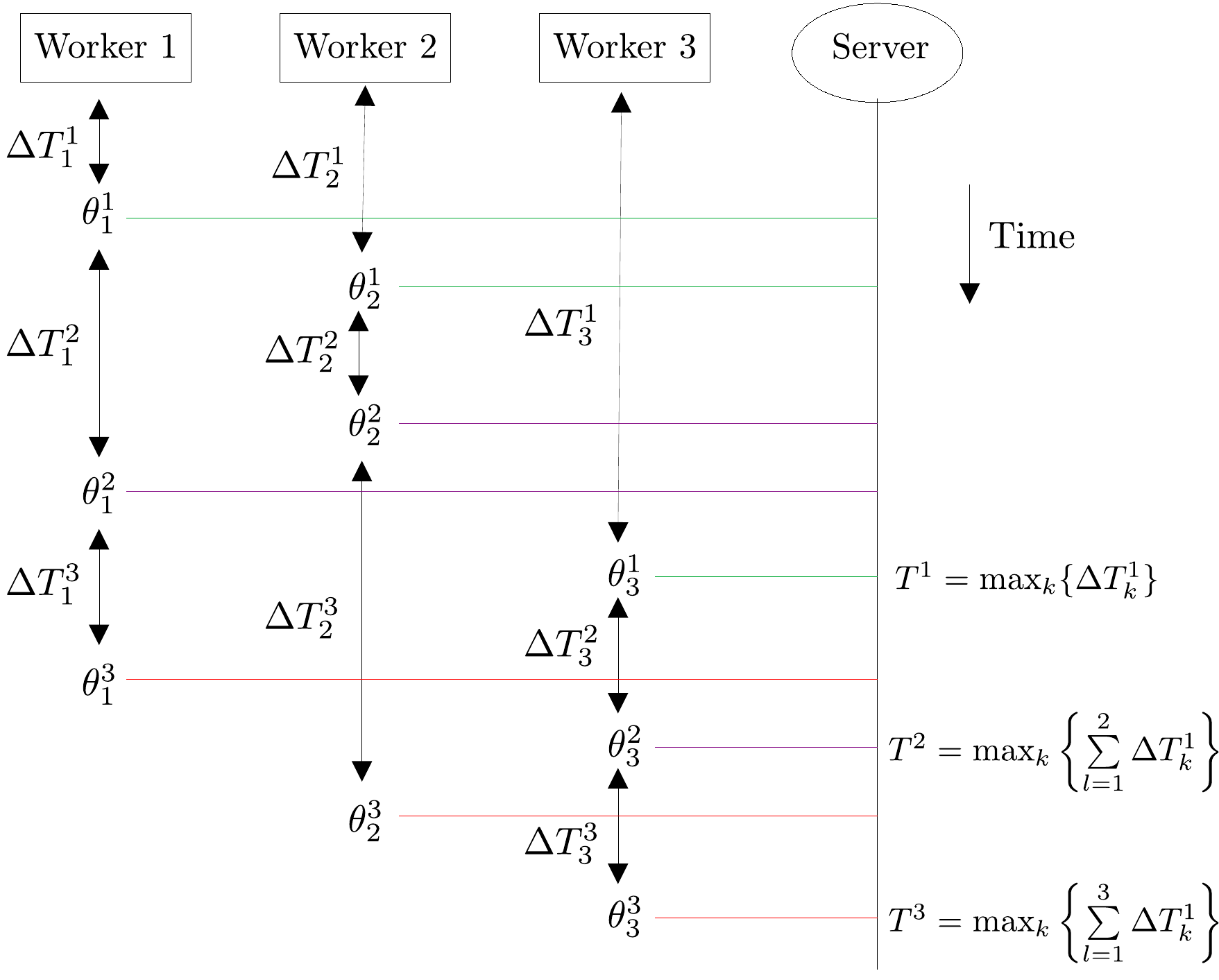} \vspace{-0.3cm}
    \caption{An illustration of computing times at $K=3$ workers and at the server. As an example, at time instant $T^2$, the server has access to $L_{1}(T^2)= 3, L_{2}(T^2)= 2, L_{3}(T^2)= 2$ local samples from the respective workers, and hence it can aggregate such sets of local samples, producing $L(T^2)=2$ global samples.}
    \label{fig:computing times}
    \vspace{-0.4cm}
\end{figure}

\section{CMC with stragglers}
\label{section GCMC}
In this section, we describe the standard CMC protocol in the context of the system under study with random computing times. The purpose of this novel formulation is to study the effect of stragglers in the CMC protocol \cite{CMC_Bayes_and_Bigdata}. Following \cite{CMC_Bayes_and_Bigdata}, we focus here on the standard case with no computing redundancy, i.e., $r=1$, and we let each data shard, $\mathcal{Z}_s$, be allocated only to worker $k=s$. Throughout this section, we accordingly simplify the notation by writing $\theta_{k}^{l}$ for the sample $\theta_{k,k}^{l}$ generated at worker $k$ for the $k$-th shard $\mathcal{Z}_k$. Note also that we have $\Theta_k^l = \{ \theta_{k,k}^{l} \} = \{ \theta_k^l\}$.

Each worker $k$ communicates a generated sample $\theta_{k}^{l}$ to the server as soon as it is produced, where index $l\in \{1,2,\cdots\}$ runs over the samples. Given the model described in Section \ref{section system model}, we denote as $L_k(t)$ the number of samples $\{\theta_{k}^l \}_{l=1}^{L_k(t)}$ received by the server up to time $t$ from worker $k$, which is given by 
\begin{align}
\label{eqn L_k(t)}
    L_k(t) &=\max \left\{ l: \sum\limits_{l^{\prime} =1}^{l}\Delta T_{k}^{l^{\prime}} \leq t \right\}.    
\end{align} 
As soon as all the $K$ local samples $\{\theta_{k}^l \}_{k=1}^{K}$ are received, a global sample $\theta^l$ can be computed at the server by aggregating the corresponding local samples $\{\theta_{k}^l \}_{k=1}^{K}$. Therefore, at time $t$, the number of global samples aggregated at the server is $L(t)= \min_{k \in [K]} L_k(t)$. 
An illustration of the computing times of each worker and the server are shown in Fig \ref{fig:computing times}, with $K=3$ workers. 
The computing time of global samples at the server is limited by the computing time of worker 3.

CMC makes the working assumption that the local samples $\theta_k^l$ are Gaussian $\mathcal{N}(\mu_k,C_k)$ with mean $\mu_k$ and covariance $C_k$.  Under this condition, which is practically and approximately valid only as $N/K \rightarrow \infty$, the optimal aggregation function is 
\vspace{-0.2cm} \cite{CMC_Bayes_and_Bigdata}
\begin{equation}
\label{eqn CMC aggregation}
    \theta^l = \sum\limits_{k=1}^{K} W_k \theta_{k}^{l},
\end{equation} \vspace{-0.4cm} 
with weight matrices 
\begin{equation}
\label{eqn weight matrix in GCMC}
W_k= \left(\sum_{k=1}^{K} C_k^{-1}\right)^{-1} C_k^{-1}.
\end{equation}

The weight matrices in (\ref{eqn weight matrix in GCMC}) are not directly computable, since the parameters $\{\mu_k,C_k\}_{k=1}^{K}$ are unknown. However, at time $t$, the server can estimate the mean $\mu_k(t)$ and covariance $C_k(t)$ of the subposterior $\Tilde{p}(\theta | \mathcal{Z}_k)$ by using the $L_k(t)$ samples $\{\theta_{k}^l \}_{l=1}^{L_k(t)}$ received from the worker $k$ up to time $t$ as
\vspace{-0.2cm}
\begin{subequations}
\label{eqn mu_k and C_k estimate}
\begin{align}
    \hspace{-0.2cm} \hat{\mu}_k(t) &=\frac{1}{L_k(t)} \sum\limits_{l=1}^{L_k(t)}\theta_k^{l}, \text{   and}\\
    \label{eqn C_k estimate}
    \hspace{-0.2cm} \hat{C}_k(t) &= \sigma^2 I + \frac{1}{L_k(t)} \sum\limits_{l=1}^{L_k(t)} (\theta_{k}^{l} - \hat{\mu}_k(t)) \left(\theta_{k}^{l} - \hat{\mu}_k(t)\right)^T, 
    \vspace{-0.1cm}
\end{align}
\end{subequations} 
respectively, where $\sigma ^2$ is a regularization parameter. 
Using these estimates CMC
approximates the weight matrices in (\ref{eqn weight matrix in GCMC}) and obtain an estimate of the global sample $\theta^l$, for each $l\in [L(t)]$, at time $t$, using (\ref{eqn CMC aggregation}), as 
\vspace{-0.2cm}
\begin{equation}
\label{eqn Gaussian CMC aggregation}
    \theta^l (t) = \sum\limits_{k=1}^{K} \left(\sum_{k=1}^{K} (\hat{C}_k(t))^{-1}\right)^{-1} (\hat{C}_k(t))^{-1} ~ \theta_{k}^{l}.
\end{equation}

At any time $T^l$ at which the sever has received a new set of local samples $\{\theta_k^l\}_{k\in [K]}$, the server computes all global samples $\{\theta^{l^{\prime}}(T^l)\}_{l^{\prime}=1}^{l}$ using (\ref{eqn Gaussian CMC aggregation}), with the updated estimates of covariance matrices, $\{\hat{C}_k(T^l)\}_{k\in [K]}$, calculated using (\ref{eqn C_k estimate}).
CMC is summarized in Algorithm \ref{algorithm GCMC}.


\begin{algorithm}[htbp]
\caption{CMC with straggling workers}
\label{algorithm GCMC}
\begin{algorithmic}[1]
\State \textbf{Input:} Number of workers $K$, data shards $\{\mathcal{Z}_k\}_{k=1}^{K}$

\State \textbf{Data allocation:} 
\For{each $k\in [K]$}
\State Allocate shard $\mathcal{Z}_k$ to worker $k$
\EndFor

\State \textbf{At worker $k$:} set $l=1$

\Repeat{}
\State \hspace{0.5cm} compute $l$-th sample $\theta_k^l \sim \tilde{p}(\theta| \mathcal{Z}_k) $ \State \hspace{0.5cm} when completed, i.e., at time $\sum_{l^\prime=1}^{l}\Delta T_k^{l^\prime}$, send sample $\theta_{k}^{l}$ to the server
\EndRepeat

\State \textbf{At the server:} set $l=1$
\Repeat{}
\State when receiving all $l$-th local samples $\{\theta_k^l\}_{k\in [K]}$, i.e., at time $T^l = \max\limits_{k \in [K]} ( \sum\limits_{l^\prime=1}^{l}\Delta T_k^{l^\prime} )$, compute the covariance matrices $\{\hat{C}_k(T^l)\}_{k\in [K]}$ using (\ref{eqn mu_k and C_k estimate}), 
\vspace{0.2cm}

\State  for each $l^\prime \in [l]$ compute $\theta^{l^\prime}(T^l)$ using (\ref{eqn Gaussian CMC aggregation}) with covariance matrices $\{\hat{C}_k(T^l)\}_{k\in [K]}$

\EndRepeat

\end{algorithmic}
\vspace{-0.1cm}

\end{algorithm}


\section{Group-based CMC (G-CMC)}
\label{section group-based GCMC}

\vspace{-0.2cm}

In this section, we propose a protocol named \textit{G-CMC} that aims at leveraging the redundancy in data allocation to mitigate the effect of stragglers on the performance of CMC.
The approach clusters all the workers into groups, similar to \cite{LAGC,Grouping_in_distributed_Gradient_descent,Grouping_in_coded_caching,Grouping_in_PetrosElia_codedcaching}, and allocates the same set of $r$ shards to all the workers in a group. In this way, in order to generate the $l$-th global sample $\theta^l$, the server must only wait to receive one batch of $r$ local samples from the fastest worker in each group.

To elaborate, G-CMC partitions the set of $K$ shards, $\{\mathcal{Z}_s\}_{s\in[K]}$, into $G$ disjoint groups $\{\mathsf{Z}_g\}_{g \in [G]}$ each having $r$ shards, and the set of workers, $[K]$, into $G$ disjoint groups $\{\mathsf{K}_g\}_{g\in[G]}$ each having $r$ workers. Accordingly, we have $K=Gr$. For any $g\in [G]$, the group of shards $\mathsf{Z}_g$ is allocated exclusively to all the workers in group $\mathsf{K}_g$, i.e., $\mathcal{Z}_{\mathcal{S}_k} = \mathsf{Z}_g$ for all $k\in \mathsf{K}_g$. Therefore, each data shard is available at $r$ workers, and each worker has access to exactly $r$ shards. For $r=1$, we get $K=G$, i.e., each group has exactly one user, and G-CMC is equivalent to CMC (see Section \ref{section GCMC}).

The main idea underlying G-CMC is to treat each group as a ``super-worker'', and apply the CMC protocol presented in Section \ref{section GCMC} accross the $G$ ``super-workers''. The $l$-th batch of $r$ local samples received from the $g$-th group $\mathsf{K}_g$ is $\Theta_{\mathsf{K}_g}^{l}= \{ \theta_{\mathsf{K}_g ,s}^l \sim \tilde{p}(\theta |\mathcal{Z}_s) \}_{s: ~ \mathcal{Z}_s \in \mathsf{Z}_g}$. We denote as $L_{\mathsf{K}_g}(t) = \sum_{k\in \mathsf{K}_g} L_{k}(t)$, for $g\in [G]$, the number of batches of samples received from the group $\mathsf{K}_g$ up to time $t$, where $L_k(t)$ is the number (\ref{eqn L_k(t)}) of batches of samples generated by the user $k\in \mathsf{K}_g$ up to time $t$. The $L_{\mathsf{K}_g(t)}$ batches of samples $\{ \Theta_{\mathsf{K}_g}^l \}_{l \in [L_{\mathsf{K}_g(t)}]}$ are received at the server from the group $\mathsf{K}_g$ at time instants $\{T_{\mathsf{K}_g}^l\}_{l\in [L_{\mathsf{K}_g(t)}]}$ respectively, where $T_{\mathsf{K}_g}^l$ is the $l$-th order statistic, i.e., the $l$-th smallest value, of the variables 
$\{ \{ \sum_{l^{\prime}=1}^{i} \Delta T_{k}^{l^{\prime}} \}_{i \in [L_{k}(t)]} \}_{k\in \mathsf{K}_g}$.

As soon as all the $l$-th batches of samples, $\{\Theta_{\mathsf{K}_g}^l\}_{g\in [G]}$, with each batch having $r$ local samples are received at the server from all the $G$ groups $\{\mathsf{K}_g\}_{g\in [G]}$, a global sample $\theta^l$ is computed at the server by aggregating the corresponding $K=Gr$ local samples $\bigcup_{g\in [G]} \Theta_{\mathsf{K}_g}^l$, and hence G-CMC is resilient to $G(r-1)$ stragglers, as long as $K$ is a multiple of $r$ (see also Section \ref{section experiments}). Therefore, at time $t$, the number of global samples aggregated at the server is $L(t)= \min_{g\in [G]} L_{\mathsf{K}_g}(t)$. To this end, the server estimates the mean $\mu_s(t)$ and covariance $C_s(t)$ of the subposterior $\Tilde{p}(\theta | \mathcal{Z}_s)$, for $\mathcal{Z}_s\in \mathsf{Z}_g$, by using the $L_{\mathsf{K}_g}(t)$ samples $\{\theta_{\mathsf{K}_g,s}^l \in \Theta_{\mathsf{K}_g}^l \}_{l\in [L_{\mathsf{K}_g}(t)]}$ received from the group $\mathsf{K}_g$ up to time $t$ as

\vspace{-0.5cm}

\footnotesize
\begin{subequations}
\label{eqn mu_k and C_k estimate in G-GCMC}
\begin{align}
    \hspace{-0.3cm} \hat{\mu}_s(t) &=\frac{1}{L_{\mathsf{K}_g}(t)} \sum\limits_{l=1}^{L_{\mathsf{K}_g}(t)}\theta_{\mathsf{K}_g,s}^{l}, \text{ and } \\
    \label{eqn C_k estimate in G-GCMC}
    \hspace{-0.3cm} \hat{C}_s(t) &= \sigma^2 I + \frac{1}{L_{\mathsf{K}_g}(t)} \sum\limits_{l=1}^{L_{\mathsf{K}_g}(t)} (\theta_{\mathsf{K}_g,s}^{l} - \hat{\mu}_s(t)) (\theta_{\mathsf{K}_g,s}^{l} - \hat{\mu}_s(t))^T,
\end{align}
\end{subequations} \normalsize

\noindent respectively, where $\sigma ^2$ is a regularization parameter. 
Using these estimates, the weight matrices in (\ref{eqn weight matrix in GCMC}) are approximated to obtain the global sample $\theta^l$, for each $l\in [L(t)]$, at time $t$, using (\ref{eqn CMC aggregation}), as  \small
\begin{equation}
\label{eqn Gaussian CMC aggregation in G-GCMC}
    \theta^l (t) = \sum\limits_{s=1}^{K} \left(\sum_{s=1}^{K} (\hat{C}_s(t))^{-1}\right)^{-1} (\hat{C}_s(t))^{-1} ~ \theta_{\mathsf{K}_g,s}^{l}.
\end{equation} \normalsize
At any time $T^l$ at which the sever has received a new set of batches of samples $\{\Theta_{\mathsf{K}_g}^l\}_{g\in [G]}$, the server computes all global samples $\{\theta^{l^\prime}(T^l)\}_{l^\prime=1}^{l}$ using (\ref{eqn Gaussian CMC aggregation in G-GCMC}) with the updated estimates of covariance matrices, $\{\hat{C}_s(T^l)\}_{s\in [K]}$, calculated using (\ref{eqn C_k estimate in G-GCMC}).
G-CMC is summarized in Algorithm \ref{algorithm Group-based GCMC}.


\begin{algorithm}
\caption{Group-based CMC (G-CMC)}
\label{algorithm Group-based GCMC}
\begin{algorithmic}[1]
\State \textbf{Input:} Partition the set of workers $[K]$ into $G$ groups $\{\mathsf{K}_g\}_{g\in [G]}$ each having $r$ workers, Partition the set of shards $\{\mathcal{Z}_s\}_{s=1}^{K}$ into $G$ groups $\{\mathsf{Z}_g\}_{g\in [G]}$ each having $r$ shards, where $r=\frac{K}{G}$ is the redundancy parameter

\State \textbf{Data allocation:} 
\For{each $g\in [G]$}
\State allocate all $r$ shards in the group $\mathsf{Z}_g$ to all $r$ workers in group $\mathsf{K}_g$, i.e., $\mathcal{Z}_{\mathcal{S}_k} = \mathsf{Z}_g$ for all $k\in \mathsf{K}_g$
\EndFor

\State \textbf{At the group of workers $\mathsf{K}_g$:} set $l=1$
\Repeat{}
\State $l$-th batch of $r$ local samples  $\Theta_{\mathsf{K}_g}^l = \{\theta_{\mathsf{K}_g,s}^l \sim \tilde{p}(\theta| \mathcal{Z}_s) \}_{\mathcal{Z}_s\in \mathsf{Z}_g} $, is computed at any of the worker in the group $\mathsf{K}_g$  
\vspace{0.1cm}
\State when completed, i.e., at time $T_g^l$, send $\Theta_{\mathsf{K}_g}^{l}$ to the server, where $T_g^l$ denote the $l$-th
order statistic of the variables 
$\{ \{ \sum_{l^{\prime}=1}^{l} \Delta T_{k}^{l^{\prime}} \}_{l \in [L_{k}(t)]} \}_{k\in \mathsf{K}_g}$

\EndRepeat

\State \textbf{At the server:} set $l=1$
\Repeat{}

\State when receiving all the $l$-th batches of samples $\{\Theta_{\mathsf{K}_g}^l\}_{g\in [G]}$, at time $T^l=\max_{g \in [G]}  T_{g}^{l} $, compute the covariance matrices  $\{\hat{C}_s(T^l)\}_{s\in [K]}$ using (\ref{eqn mu_k and C_k estimate in G-GCMC})

\State for each $l^\prime \in [l]$ compute $\theta^{l^\prime}(T^l)$ using (\ref{eqn Gaussian CMC aggregation}) with covariance matrices $\{\hat{C}_s(T^l)\}_{s\in [K]}$

\EndRepeat

\end{algorithmic}
\end{algorithm}


\vspace{-0.3cm}

\section{Coded CMC (C-CMC)}
\label{section CGCMC}


In this section, we introduce C-CMC. To start, we fix a $K \times K$ encoding matrix $B$ and a $F \times K$ decoding matrix $A$ that define a \textit{gradient coding} scheme \cite{GC_Rashish_Tandon} robust to $r-1$ stragglers, with $F=\binom{K}{K-r+1}$. Accordingly, matrices $A$ and $B$ satisfy the equality $AB=1$, where $1$ is the ${F\times K}$ all-$1$ matrix. The shards are allocated to the workers according to the non-zero entries of the encoding matrix $B$, i.e., a shard $\mathcal{Z}_s$ is allocated to the worker $k$ if the $(k,s)$-th entry of $B$ is not equal to zero. The row weight and column weight of $B$ are all equal to $r$, accounting for the facts that each shard is available to $r$ workers and that each worker has access to exactly $r$ shards.

We assume that workers share common randomness, i.e., common random seeds, so that two workers $k$ and $k^\prime$ assigned the same shard $\mathcal{Z}_s$ produce the same $l$-th sample $\theta_{k,s}^{l} = \theta_{k^\prime,s}^{l}$. Accordingly, we henceforth write $\theta_{k,s}^{l} = \theta_{k^\prime,s}^{l} = \theta_{s}^{l}$. Note that common randomness is a requirement for C-CMC and not for G-CMC, which assumes the independence of the samples $\{ \theta_{k,s}^{l} \}_{k: ~ s\in \mathcal{S}_k}$ produced by all workers that are assigned the same shard $\mathcal{Z}_s$ (see Section \ref{section group-based GCMC}). 

Worker $k$ estimates the
mean $\mu_{k,s}(t)$ and covariance $C_{k,s}(t)$ of the subposterior $\tilde{p}(\theta|\mathcal{Z}_s)$, with $s\in \mathcal{S}_k$, by using $L_k(t)$ batches of samples computed by it up to time $t$ as 

\vspace{-0.4cm}

\small
\begin{subequations}
\label{eqn mu_k and C_k estimate in C-GCMC}
\begin{align}
    \hspace{-0.1cm} \hat{\mu}_{k,s}(t) &=\frac{1}{L_k(t)} \sum\limits_{l^\prime=1}^{L_k(t)}\theta_{s}^{l^\prime}, \text{ and}\\
    \label{eqn C_k estimate in C-GCMC}
    \hspace{-0.1cm} \hat{C}_{k,s}(t) &= \sigma^2 I + \frac{1}{L_k(t)} \sum\limits_{l^\prime =1}^{L_k(t)} (\theta_{s}^{l^\prime} - \hat{\mu}_{k,s}(t)) (\theta_{s}^{l^\prime} - \hat{\mu}_{k,s}(t))^T 
\end{align}
\end{subequations} \normalsize
respectively, where $\sigma ^2$ is a regularization parameter. This is in contrast to CMC and G-CMC, where the mean vector and covariance matrix of the subposterior are estimated at the server.

At time $t= T_k^l = \sum_{l^{\prime}=1}^{l}\Delta T_{k}^{l^\prime}$, worker $k$ computes the $l$-th batch of samples $\Theta_k^l = \{\theta_s^l\}_{s\in [\mathcal{S}_k]}$. 
Then, it updates the covariance matrices 
$\hat{C}_{k,s}(t)$ using (\ref{eqn mu_k and C_k estimate in C-GCMC}) for all $s\in \mathcal{S}_k$. Given the assumption of common random seeds described above, the estimates $\hat{C}_{k,s}(t)$ and $\hat{C}_{k^\prime ,s}(t^\prime)$ evaluated at any two workers $k$ and $k^\prime$, with $s\in \mathcal{S}_k$, $s\in \mathcal{S}_{k^\prime}$ and $L_k(t)=L_{k^\prime}(t^\prime)$, are equal.

Let $b_k$ be the $1\times K$ row vector of the encoding matrix $B$ corresponding to the worker $k$. Let $\Theta_k^l = \{\theta_{s_i}^l \}_{i\in [r]}$, with $s_i \in \mathcal{S}_k$ for each $i
\in [r]$, be the $l$-th batch of $r$ samples computed at the worker $k$.
Each sample $\theta_{s_i}^l$ is pre-processed as $(\hat{C}_{k ,s_i}(T_k^l))^{-1} \theta_{s_i}^l$ and the resulting processed samples are encoded using the encoding vector $b_k$ as
\begin{align}
        \label{eqn transmission in C-GCMC}
        \hspace{-0.2cm}\tilde{\theta}_k^l &=  [(\hat{C}_{k,s_1}(T_k^l))^{-1} \theta_{s_1}^{l} ~~ \cdots ~~
        (\hat{C}_{k,s_r}(T_k^l))^{-1} \theta_{s_r}^l ] (\tilde{b}_k)^T,
\end{align}
with $\tilde{b}_k$ being the $1\times r$ row vector given as $[b_k(s_1) ~~ b_k(s_2)~~ \cdots ~~ b_k(s_r)]$, where $b_k(s_i)$ is the $s_i$-th element of $b_k$.

The server waits until it receives transmissions corresponding to $l$-th samples from at least $K-r+1$ workers, and proceeds to decoding to finally compute the $l$-th global sample $\theta^l$. This occurs at time $T^l$ equal to the $(K-r+1)$-th order statistic, i.e., the $(K-r+1)$-th smallest value, of the variables $\{ T_k^l \}_{k\in [K]}$, and hence C-CMC is resilient to $r-1$ stragglers.
Let the set of $K-r+1$ non-stragglers for the $l$-th sample be indexed by an integer $j\in [\binom{K}{K-r+1}]$ and $a_j$ be the corresponding $1\times K$ row vector of the decoding matrix $A$. Let $\mathcal{K}_j^l = \{k_1^l,k_2^l,\cdots,k_{K-r+1}^l\} \subseteq [K]$ be the corresponding subset of $K-r+1$ non-stragglers. The server decodes the sum of the processed $l$-th samples, by using the transmissions from the workers in $\mathcal{K}_j^l$, as \vspace{-0.2cm}
\begin{align}
    \label{eqn sum of weighted samples in C-GCMC}
        [\tilde{\theta}_{k_1^l}^l ~~ \tilde{\theta}_{k_2^l}^l ~~\cdots ~~ \tilde{\theta}_{k_{K-r+1}^l}^l ] (\tilde{a}_j)^T = \sum_{s=1}^{K}(\hat{C}_{s}^l)^{-1} \theta_{s}^l = \phi^l.
\end{align} 
with $\tilde{a}_j$ being the $1\times (K-r+1)$ row vector given by $[a_j(k_1^l) ~ a_j(k_2^l) ~ \cdots ~ a_j(k_{K-r+1}^l)]$, where $a_j(k_i^l)$ is the $k_i^l$-th element of the $1\times K$ row vector $a_j$ for any $i\in [K-r+1]$. The matrix $\hat{C}_{s}^l$ is the empirical covariance matrix of the subposterior $\tilde{p}(\theta|\mathcal{Z}_s)$ computed using the first $l$ samples, computed at any worker in $\mathcal{K}_j^l$, and is equal to $ \hat{C}_{k_i^l,s}^l(T_{k_i^l}^l)$ for any $i\in [K-r+1]$.

In order to compute the $l$-th global sample $\theta^l$ in (\ref{eqn Gaussian CMC aggregation}) the server has to pre-multiply the decoded sample $\phi^l$ in (\ref{eqn sum of weighted samples in C-GCMC}) with the matrix $(\sum_{s=1}^{K}(\hat{C}_s^l)^{-1})^{-1} $. We propose that the server estimates the matrix $\sum_{s=1}^{K}(\hat{C}_s^l)^{-1}$ as the empirical covariance of the $l$ decoded samples $\phi^{l^\prime}$ for $l^\prime \in [l]$, i.e., as the matrix \vspace{-0.2cm}
\begin{align}
\label{eqn estimating the empirical covariance in C-GCMC}
    \hat{D}^l = \frac{1}{l} \sum_{l^\prime =1}^{l} (\phi^{l^\prime} - \overline{\phi})(\phi^{l^\prime} - \overline{\phi})^T \approx \sum_{s=1}^{K}C_s^{-1}
\end{align} \vspace{-0.2cm}

\noindent with $\overline{\phi}= \frac{1}{l} \sum_{l^\prime =1}^{l} \phi^{l^\prime}$. The approximate equality in (\ref{eqn estimating the empirical covariance in C-GCMC}) can be seen to be exact when $l\rightarrow \infty$. Accordingly the final global sample is $\theta^l = (\sigma^2 I+ \hat{D}^l)^{-1} \phi^l$, where $\sigma^2$ is a regularization parameter. 
The overall algorithm is summarized in Algorithm \ref{algorithm CGCMC}. The rationale for the proposed estimate (\ref{eqn estimating the empirical covariance in C-GCMC}) is provided in the Appendix. 

\vspace{-0.2cm}


\begin{algorithm}
\caption{Coded CMC (C-CMC)}
\label{algorithm CGCMC}
\begin{algorithmic}[1]
\State \textbf{Input:} Number of workers $K$, Maximum number of stragglers $r-1$, Data shards $\{\mathcal{Z}_s\}_{s=1}^{K}$
\State Consider two matrices $A,B$ such that it forms a GC scheme robust to $r-1$ stragglers \cite{GC_Rashish_Tandon}

\State \textbf{Data allocation:}
    \For{each $k,s \in [K]$}
        \If{$B(k,s) \neq 0$}
            \State Allocate shard $\mathcal{Z}_s$ to worker $k$
        \EndIf
    \EndFor

\State \textbf{At worker $k$:} set $l=1$

\Repeat{}
\State computes the $l$-th batch of samples $\Theta_k^l = \{\theta_s^l\}_{s\in [\mathcal{S}_k]}$

\State when completed, i.e., at time $T_k^l = \sum\limits_{l^{\prime}=1}^{l}\Delta T_{k}^{l^\prime}$, worker $k$ estimates the covariance matrices $\{\hat{C}_{s}^l\}_{s\in \mathcal{S}_k}$ of the subposteriors $\{\tilde{p}(\theta |\mathcal{Z}_s)\}_{s\in \mathcal{S}_k}$ using the $l$ batches of samples $\{\Theta_k^{l^\prime}\}_{l^\prime \in [l]}$ computed at the worker $k$.  

\vspace{0.2cm}

\State worker $k$ transmits $\tilde{\theta}_k^l $ using (\ref{eqn transmission in C-GCMC})


\EndRepeat

\State \textbf{At the server:} set $l=1$
\Repeat{}
\State using the transmissions from $K-r+1$ non-stragglers, compute $\sum_{s=1}^{K}(\hat{C}_{s}^l)^{-1} \theta_{s}^l$ using (\ref{eqn sum of weighted samples in C-GCMC}) at time $T^l$ being equal to $(K-r+1)$-th order statistic of the variables $\{ T_k^l \}_{k\in [K]}$
\vspace{0.2cm}

\State Estimate $\sum_{s=1}^{K}(\hat{C}_s^l)^{-1}$ using (\ref{eqn estimating the empirical covariance in C-GCMC}) and compute global sample $\theta^l$ using (\ref{eqn Gaussian CMC aggregation}).

\EndRepeat

\end{algorithmic}
\end{algorithm}




\vspace{-0.3cm}

\section{Experiments}
\label{section experiments}
In this section, we evaluate the performance of the considered CMC schemes in the presence of stragglers. We present two experiments on distributed computing systems with $K=5$ and $K=40$ workers\footnote{We have carried out all experiments on a laptop with i7 processor and $16$ GB RAM by using MATLAB. Code is available at $<$https://github.com/kclip/Straggler-resilient-CMC$>$.}. In both experiments, we assume that the subposterior $\Tilde{p}(\theta | \mathcal{Z}_s)$ for each shard $\mathcal{Z}_s$ where $s\in [K]$ to be a $5$-dimensional multivariate Gaussian distribution $\mathcal{N}(0,C_s)$ with a symmetric Toeplitz covariance matrix $C_s$ using first column $[1 ~\rho_s ~\rho_s^2 ~\rho_s^3 ~\rho_s^4]^T$ with $\rho_s = (s-1)/K$ \cite{Dongzhu}. As in \cite{Variational_CMC,Dongzhu}, we calculate the error in (\ref{eqn error}) for multiple test functions, with each test function $f_{i,j}(\theta) = \theta[i]\theta[j]$ being an element in the outer product matrix $\theta \theta^T$ for $i,j\in [d]$, and average them to obtain the final error. We consider Pareto distribution with $\eta = 0.1$ and shape parameter $\beta = 1.2$, which corresponds to scale parameter $r/60$ and mean $0.1r$ (see Section \ref{section system model}), for the computing time $\Delta T_{k}^{l}$ at the workers (see, e.g., \cite{LAGC}).


\begin{figure}[t]
    \centering
    \includegraphics[height=2.6in]{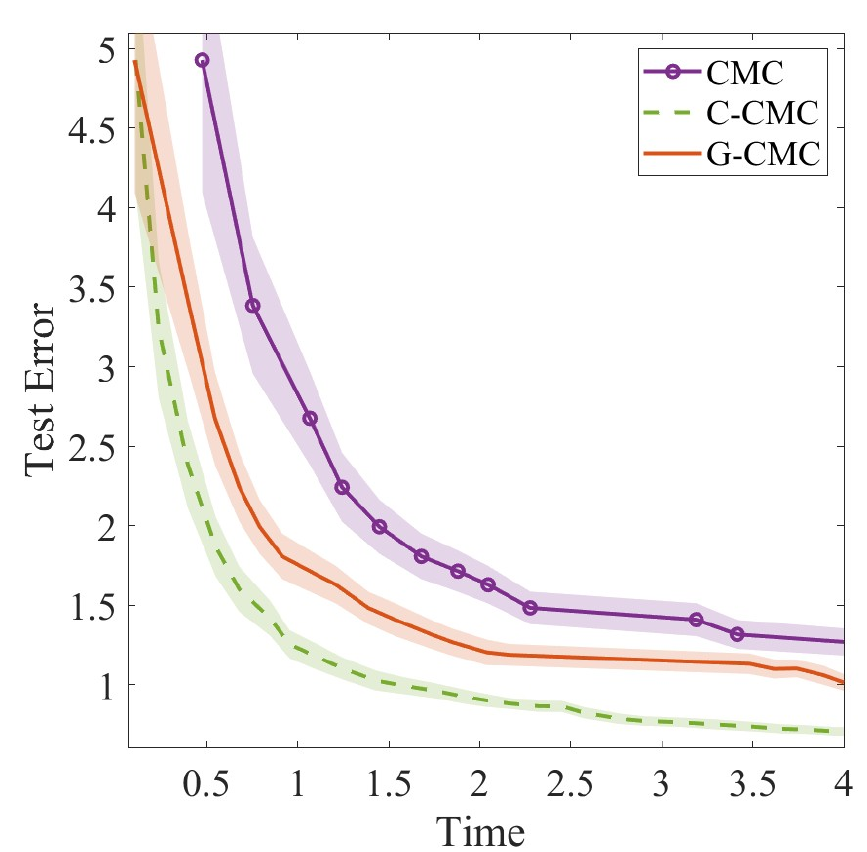} 
    \vspace{-0.3CM}
    \caption{Average test error versus time with $K=5$ workers and $r=2$. C-CMC is resilient to $1$ straggler. As $K/r$ is not an integer, we apply G-CMC on a group of $4$ workers, with the last group having one worker. As a result, G-CMC is not resilient to stragglers, from the last group.}
    \label{fig:error vs time with K=5 workers} 
    \vspace{-0.2cm}
\end{figure}


\begin{figure}[t]
    \vspace{-0.1cm}
    \centering
    \includegraphics[height=2.6in]{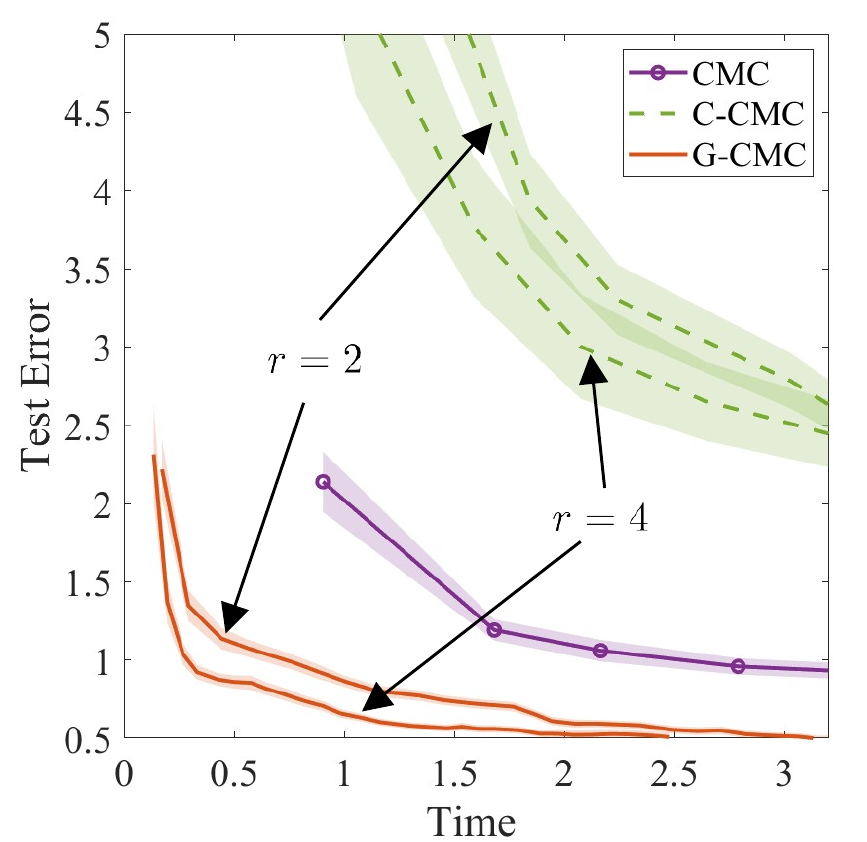} 
    \vspace{-0.2cm}
    \caption{Average test error versus time with $K=40$ workers. For $r=2$, C-CMC and G-CMC are resilient to $1$ and $20$ stragglers respectively. For $r=4$, C-CMC and G-CMC are resilient to $3$ and $30$ stragglers respectively.}
    \label{fig:error vs time with K=40 workers}  
    \vspace{-0.2cm}
\end{figure}




In Fig. \ref{fig:error vs time with K=5 workers} and Fig. \ref{fig:error vs time with K=40 workers}, we plot the test error averaged over $50$ random realizations of computing times, as a function of time, at $K=5$ and $K=40$ workers respectively, following a Pareto distribution with $\eta =0.1$ and shape parameter $\beta = 1.2$, which corresponds to scale parameter $r/60$, for the Pareto distribution with mean $0.1r$. The curve represents the average error, and shaded region represents the error bars corresponding to $0.15$ times the standards deviation of the error across the realizations. We plot the curves for CMC, C-CMC and G-CMC. For $K=5$, as $K/r$ is not an integer, G-CMC is applied to three groups with one of the groups having a single worker. If the straggler is from the group containing a single worker, this delays the computation of global sample at the server, which in turn degrades the performance of G-CMC for smaller $K$. Note that C-CMC can be directly applied as there is no such restriction on $r$ in C-CMC. From Fig. \ref{fig:error vs time with K=5 workers} and Fig. \ref{fig:error vs time with K=40 workers}, we can conclude that G-CMC is more efficient for straggler mitigation for large values of $K$, where as C-CMC is efficient for smaller values of $K$.

\section{Conclusions}
\label{section conclusion}
In this paper, we have considered, for the first time, the problem of stragglers in Consensus Monte Carlo (CMC) for distributed Bayesian learning. We studied the effect of stragglers in the standard CMC protocol, and proposed two schemes, Group-based CMC (G-CMC) and Coded CMC (C-CMC) that effectively leverage the redundancy in data allocation. 






\section*{Appendix}

In this Appendix, we detail the approximation in (\ref{eqn estimating the empirical covariance in C-GCMC}).
To this end, we write the empirical covariance of the $l$ decoded samples $\phi^{l^\prime}$, for $l^\prime \in [l]$, as \small
\begin{align}
    \nonumber
    \hat{D}^l &=  \frac{1}{l} \sum_{l^\prime =1}^{l} \left(\phi^{l^\prime}  (\phi^{l^\prime})^T \right) - \overline{\phi} (\overline{\phi})^T  \\
    \nonumber
    & = \frac{1}{l} \sum_{l^\prime = 1}^{l} \left[ \left( \sum_{s=1}^{K}(\hat{C}_{s}^{l^\prime})^{-1} \theta_{s}^{l^\prime}  \right) \left( \sum_{m=1}^{K}(\theta_{m}^{l^\prime})^T (\hat{C}_{m}^{l^\prime})^{-1}   \right) \right] \\
     \nonumber
     & ~~  - \left(\frac{1}{l} \sum_{l^\prime =1}^{l}\sum_{s=1}^{K}(\hat{C}_{s}^{l^\prime})^{-1} \theta_{s}^{l^\prime}\right)  \left(\frac{1}{l} \sum_{l^\prime =1}^{l}\sum_{m=1}^{K}(\theta_{m}^{l^\prime})^T (\hat{C}_{m}^{l^\prime})^{-1} \right) \\
     \nonumber
     & \stackrel{(a)}{\approx}  \frac{1}{l} \sum_{l^\prime = 1}^{l} \left[ \left( \sum_{s=1}^{K}C_s^{-1} \theta_{s}^{l^\prime}  \right) \left( \sum_{m=1}^{K}(\theta_{m}^{l^\prime})^T C_m^{-1}  \right) \right] \\
    \nonumber
     & ~~~    - \left(\frac{1}{l} \sum_{l^\prime =1}^{l}\sum_{s=1}^{K}C_s^{-1} \theta_{s}^{l^\prime}\right)  \left(\frac{1}{l} \sum_{l^\prime =1}^{l}\sum_{m=1}^{K}(\theta_{m}^{l^\prime})^T C_m^{-1} \right)
\end{align} \normalsize
where, the approximation in (a) is justified by the fact that, as $l\rightarrow \infty$, the covariance matrix $\hat{C}_s^l$ converges to the ground-truth covariance matrix $C_s$. Letting $\mu_s^l = \frac{1}{l} \sum_{l^\prime=1}^{l} \theta_s^{l^\prime}$ be the mean of first $l$ samples of $\tilde{p}(\theta|\mathcal{Z}_s)$, we can write $\hat{D}^l$ as

\small
\begin{align}
     \nonumber
    \hat{D}^l & \approx  \frac{1}{l} \sum_{l^\prime = 1}^{l} \left[ \left( \sum_{s=1}^{K}C_s^{-1} \theta_{s}^{l^\prime}  \right) \left( \sum_{m=1}^{K}(\theta_{m}^{l^\prime})^T C_m^{-1}  \right) \right] \\
    \nonumber
    & ~~~   - \frac{1}{l} \sum_{l^\prime = 1}^{l} \left[\left(\sum_{s=1}^{K}C_s^{-1} \mu_{s}^l\right)  \left(\sum_{m=1}^{K}(\mu_{m}^l)^T C_m^{-1} \right) \right] \\
    \nonumber
    & = \frac{1}{l} \sum_{l^\prime = 1}^{l}  \sum_{s=1}^{K} \left[ C_s^{-1} \theta_{s}^{l^\prime}    (\theta_{s}^{l^\prime})^T C_s^{-1}  -  C_s^{-1} \mu_{s}^{l}    (\mu_{s}^{l})^T C_s^{-1} \right]  \\
    \nonumber
    & ~~   + \hspace{-0.05cm}  \frac{1}{l} \hspace{-0.05cm}  \sum_{l^\prime = 1}^{l}  \sum_{\substack{s,m=1 \\ s\neq m }}^{K} \hspace{-0.1cm} \left[ C_s^{-1} \theta_{s}^{l^\prime}    (\theta_{m}^{l^\prime})^T C_m^{-1}  - C_s^{-1} \mu_{s}^{l}    (\mu_{m}^{l})^T C_m^{-1} \right] \\
    \nonumber
    & \stackrel{(b)}{\approx}  \sum_{s=1}^{K}C_s^{-1} \mathrm{E}[\theta_{s}^{l^\prime}    (\theta_{s}^{l^\prime})^T - \mu_{s}^{l}    (\mu_{s}^{l})^T] C_s^{-1}  \\
    \nonumber
    & ~~~   +  \sum_{\substack{s,m=1 \\ s\neq m }}^{K}C_s^{-1} \mathrm{E} [ \theta_{s}^{l^\prime}    (\theta_{m}^{l^\prime})^T - \mu_{s}^{l} (\mu_{m}^{l})^T] C_m^{-1}
\end{align}
\begin{align}
    \nonumber
    & = \sum_{s=1}^{K}C_s^{-1} C_s C_s^{-1}    +   \sum_{\substack{s,m=1 \\ s\neq m }}^{K}C_s^{-1} (0) C_m^{-1} = \sum_{s=1}^{K}C_s^{-1},
\end{align}
\normalsize
\noindent where the approximation $(b)$ is exact as $l\rightarrow \infty$.

\bibliographystyle{IEEEtran}
\bibliography{IEEEabrv,cite.bib}

\end{document}